\documentclass[runningheads]{llncs}
\usepackage{graphicx}

\usepackage{tikz}
\usepackage{comment}
\usepackage{amsmath,amssymb} 
\usepackage{color}
\usepackage{graphicx}
\usepackage{booktabs}
\usepackage{multirow}
\usepackage{xcolor}
\usepackage{float}
\usepackage{multirow}
\usepackage{subcaption}
\usepackage[accsupp]{axessibility}  

\begin{document}
\pagestyle{headings}
\mainmatter
\def\ECCVSubNumber{4486}  

\title{InsCon: Instance Consistency Feature Representation via Self-Supervised Learning} 




\titlerunning{Instance Consistency Feature Representation via Self-Supervised Learning}
%
%
\authorrunning{Junwei Yang et al.}
%
\author{Junwei Yang\textsuperscript\dag, Ke Zhang\textsuperscript\dag\textsuperscript{*}, Zhaolin Cui, Jinming Su, Junfeng Luo\textsuperscript{*},\\ and Xiaolin Wei} 

\institute{Meituan \\
\email{\{yangjunwei03,zhangke21,cuizhaolin,sujinming,\\luojunfeng,weixiaolin02\}@meituan.com}}

\renewcommand{\thefootnote}{\fnsymbol{footnote}} 
\footnotetext[4]{These authors contributed equally to this work.} 
\footnotetext[1]{Co-corresponding author.}

\maketitle

\begin{abstract}
Feature representation via self-supervised learning has reached remarkable success in image-level contrastive learning, which brings impressive performances on image classification tasks. While image-level feature representation mainly focuses on contrastive learning in single instance, it ignores the objective differences between pretext and downstream prediction tasks such as object detection and instance segmentation. In order to fully unleash the power of feature representation on downstream prediction tasks, we propose a new end-to-end self-supervised framework called InsCon, which is devoted to capturing multi-instance information and extracting cell-instance features for object recognition and localization. On the one hand, InsCon builds a targeted learning paradigm that applies multi-instance images as input, aligning the learned feature between corresponding instance views, which makes it more appropriate for multi-instance recognition tasks. On the other hand, InsCon introduces the pull and push of cell-instance, which utilizes cell consistency to enhance fine-grained feature representation for precise boundary localization. As a result, InsCon learns multi-instance consistency on semantic feature representation and cell-instance consistency on spatial feature representation. Experiments demonstrate the method we proposed surpasses MoCo v2 by 1.1\% {AP$^{bb}$} on COCO object detection and 1.0\% {AP$^{mk}$} on COCO instance segmentation using Mask R-CNN R50-FPN network structure with 90k iterations, 2.1\% {AP$^{bb}$} on PASCAL VOC objection detection using Faster R-CNN R50-C4 network structure with 24k iterations.
\keywords{Feature representation \and cell-instance \and multi-instance}
\end{abstract}

\section{Introduction}
\label{sec:intro}

Over the past few years, supervised learning has established a mature training paradigm in computer vision tasks. Generally, models are pre-trained on large-scale labeled datasets like ImageNet\cite{deng2009imagenet} and then transferred to targeted visual tasks for fine-tuning. This pattern heavily relies on data annotation to get better performance. However annotated images occupy only a small part of overall images in the real world. When transferring to a certain scenario with little annotated data, models may suffer from severe performance degradation.
\begin{figure*}[ht]
\centering
\includegraphics[width=1.0\linewidth]{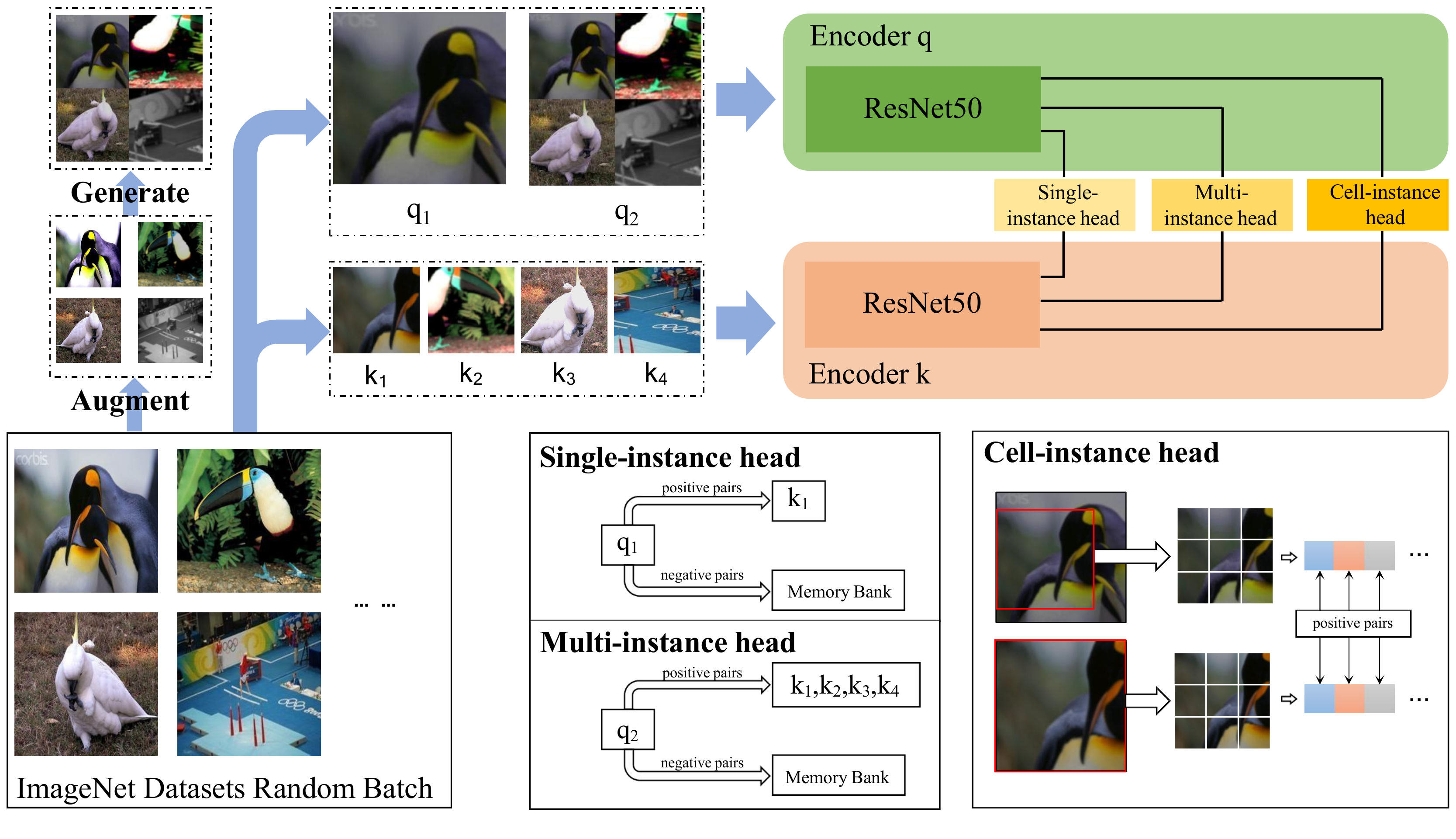}
\caption{For query encoder, InsCon takes q$_{1}$ and q$_{2}$ as input. q$_{1}$ is the original view from MoCo v2, and q$_{2}$ is composed by four different augmented views coming from the same batch. For the key encoder, InsCon takes randomly augmented views(k$_{1}$, k$_{2}$, k$_{3}$, k$_{4}$) from original images as the input. The output of backbone(ResNet50) will send into three parallel heads. Single-instance head represents the original head used in MoCo v2 which constructs instance consistency of q$_{1}$ and \{k$_{1}$, memory bank\} for contrastive learning. Multi-instance head will learn features of multiple instances by constructing the instance consistency of q$_{2}$ and \{k$_{1}$, k$_{2}$, k$_{3}$, k$_{4}$, memory bank\}. Cell-instance head will learn fine-grained features from cells, and each cell represents a small feature region.}
\label{fig:alg structure}
\end{figure*}

\vspace{-0.5cm}
Recently, self-supervised learning has aroused lots of attention, which aims to build an excellent feature extractor from large unlabeled images. Some researches\cite{he2020momentum,chen2020improved,chen2020simple,grill2020bootstrap,tian2020makes,xiao2021region,xie2021detco,xie2021propagate,wang2021dense} have shown their success on image classification and downstream prediction tasks by fine-tuning the pre-trained model in self-supervised learning. A common self-supervised method based contrastive learning\cite{hadsell2006dimensionality} like MoCo v2\cite{chen2020improved} follows simple instance discrimination tasks\cite{wu2018unsupervised,xie2017aggregated,bachman2019learning} which match encoded queries and keys of same images. This paradigm is well-suited for single instance learning, especially for classification tasks. However, it brings gaps when transferring to downstream prediction tasks yet. On the one hand, images collected in the real world always cover multiple instances which brings more complex recognition scenarios. For example, some challenging tasks on COCO\cite{lin2015microsoft} and PASCAL VOC\cite{everingham2010pascal} need to recognize and locate multiple instances in a single image. On the other hand, complex recognition scenarios need to focus on fine-grained features to capture detailed information. For example, object detection task needs to locate all instances' boundaries and assign corresponding categories to them, instance segmentation needs to accomplish pixel-level category learning. Although some methods\cite{tian2020makes,xiao2021region,xie2021detco,xie2021propagate,wang2021dense} have explored the potential of self-supervised learning on downstream prediction tasks, there still remains huge exploration space. The inner mechanism of above methods tend to assign one pseudo label for one picture or one patch, which ignores the fact some challenging tasks on COCO and PASCAL VOC need
to recognize and locate multiple instances in a single image. This paper aims to design a more general and effective instance learning paradigm.

Inspired by these observations and analysis. We propose instance consistency feature representation (InsCon) via self-supervised learning as shown in Figure \ref{fig:alg structure}. InsCon is devoted to exploring consistency of instance contrastive learning which contains three instance-related learning heads. Firstly, we retain the original learning head of MoCo v2, this learning head mainly concentrates on single instance learning, in this paper, we call it single-instance head. Secondly, to alleviate the gap between pretext and downstream prediction tasks, we introduce a multi-instance head. This head takes a mixed view as query and multiple corresponding views as positive keys, which aims to construct complex tasks closer to the downstream and improve the model's ability to recognize and locate objects. Thirdly, to further improve the performance on downstream prediction tasks, we propose a cell-instance head by establishing strict cell feature correspondence mechanism, noting that each cell represents a small region mapped to the original image. This module enhances the fine-grained feature representation ability of the model. Eventually, we design a unified loss function for different granular features to construct an end-to-end learning framework.

Our main contributions can be summarized as follows:
\begin{itemize}
\normalsize
    \item We propose a multiple instance learning paradigm named multi-instance head and combine it with single-instance learning head to enhance the performance on object recognition and localization.
    \item We introduce an effective cell-instance head to form a complementary relationship with single-instance head and multi-instance head, which further improves the model's ability on fine-grained feature representation.
    \item We create an end-to-end self-supervised learning framework with a unified contrastive loss and eventually outperform state-of-the-art methods when transferring to downstream prediction tasks.
\end{itemize}

\section{Related Works}
\label{sec:rela}

\noindent\textbf{Self-supervised learning approaches for image representation.} Compared with the supervised training paradigm,  the success of self-supervised learning can be summarized as pretext tasks\cite{vincent2008extracting,pathak2016context,zhang2016colorful,zhang2017split} and contrastive loss function\cite{hadsell2006dimensionality}. Commonly, pretext tasks try to create different views from one image, these views belong to the same pseudo label, but each view is randomly augmented by several data augmentation operators such as random crop and resize, horizontal flip, color distortion\cite{howard2013some}, Gaussian blur, gray scale and so on. The design of loss function can often be independent of the pretext tasks. Contrastive loss is widely used as a classic self-supervised loss. It guides the model to learn the semantic information from the image by constructing positive and negative pairs, so as to build a strong feature representation network.

Previous works can be mainly divided into two frameworks: SimCLR \cite{chen2020simple} and MoCo \cite{he2020momentum}. MoCo builds a huge and consistent feature dictionary named memory bank, which is used to construct negative samples. MoCo v2 is further developed on the basis of MoCo, adjusting the augmentation, projection, and other parts to achieve competitive performance. SimCLR uses a combination of image augmentation and contrastive learning to construct the self-supervised learning framework. The introduction of nonlinear MLP in projection is also a major innovation, and it has been used by \cite{chen2020improved,grill2020bootstrap,wang2021dense} and so on. These methods achieved state-of-the-art results at the time. Meanwhile, there still exist other excellent works. BYOL \cite{grill2020bootstrap} based on the above research, designs an online-target network without contrastive loss, using only two different augmentations to construct positive pairs. InfoMin\cite{tian2020makes} explores the effective views in contrastive learning to reduce mutual information while keeping the task-relevant information to improve model performance.

\noindent\textbf{Visual representation for downstream prediction tasks.} Instance discrimination pretext used in MoCo v2 is well suited for image classification since it tends to express the overall feature of images. However, it lacks attention on fine-grained feature learning, which means it can not boost the performance in downstream. Therefore, some researches propose to design an extra fine-grained feature learning module to get better results on downstream prediction tasks.
VADeR\cite{pinheiro2020unsupervised} designs pixel-level contrastive learning to guide local features to learn consistency representation. DenseCL\cite{wang2021dense} proposes a dense self-supervised learning method that directly works at the level of pixels by taking into account the correspondence between pixels. ReSim\cite{xiao2021region} learns both regional representations for localization as well as semantic image-level representations. PixPro\cite{xie2021propagate} proposes PPM combined with pixcontrast module to extend instance-level learning to pixel-level learning by using cosine similarity to calculate pixel distance which determines the selection of positive and negative samples.

In self-supervised learning, the performance of the model is usually verified on specific tasks such as object detection and instance segmentation. Generally, ResNet\cite{he2016deep} model is adopted as backbone network in pre-training stage. Recently, some excellent works such as ViT\cite{dosovitskiy2020image}, Swin transformer\cite{liu2021swin} and so on adopt Transformer\cite{vaswani2017attention} as their backbone to solve computer tasks. Meanwhile, ResNet is still the widely used network architecture and most existing researches in self-supervised learning take it as backbone. Besides, the method proposed in this paper is applicable to any backbone network. Considering above factors, we adopt ResNet as backbone in our work, Faster R-CNN\cite{ren2016faster} and Mask R-CNN\cite{he2017mask} are used in fine-tuning stage, which can be easily implemented by computer vision library such as Detectron2\cite{wu2019detectron2} and MMdetection\cite{mmdetection}. They intelligently integrate the latest object detection, instance segmentation and other models. To align the network structure on downstream prediction tasks, we carefully design configuration files and present unified experimental results.

\section{Method}
\label{sec:method}

\subsection{Preliminaries}
\label{subsec:background}

The milestone of self-supervised representation learning is the emergence of MoCo v1/v2\cite{he2020momentum,chen2020improved} and SimCLR. Both of them concentrate on instance discrimination tasks, which use random augmentation operators to produce two different views from the same image. In MoCo v2, the overall pipeline can be split into two parts, feature representation and contrastive learning.

\noindent\textbf{Feature representation.} The overall network structure can be split into two parts named query encoder and key encoder. Each encoder receives a view as input and outputs a visual representation in embedding space after passing through average pooling layer and non-linear MLP layer. Parameters in the query encoder are updated by back-propagation while parameters in the key encoder are updated by the momentum update strategy.

\noindent\textbf{Contrastive learning.} Contrastive learning procedure can be seen as mining features from a large amount of unlabeled data. Different views created by the same image will be encoded separately by query encoder and key encoder and finally make up positive pairs. Meanwhile, MoCo v2 constructs a huge dynamic feature queue to store past features coming from the key encoder, these features will play roles as negative samples against query features. The contrastive loss uses InfoNCE\cite{oord2018representation}:
\begin{equation}
\mathcal{L}_{q}=-\log \frac{\exp \left(q \cdot k_{+} / \tau\right)}{\exp \left(q \cdot k_{+}\right)+\sum_{k_{-}} \exp \left(q \cdot k_{-} / \tau\right)}
\end{equation}
Where $\tau$ is a temperature hyper-parameter in \cite{wu2018unsupervised}. In general, the contrastive loss can shorten the distance of positive pairs($q$ and $k_{+}$) and push the negative pairs($q$ and $k_{-}$) far away at the same time in feature space.

\subsection{Our Pipeline}
\label{subsec:our method}

\begin{figure}[t]
    \centering
    \includegraphics[width=0.6\linewidth]{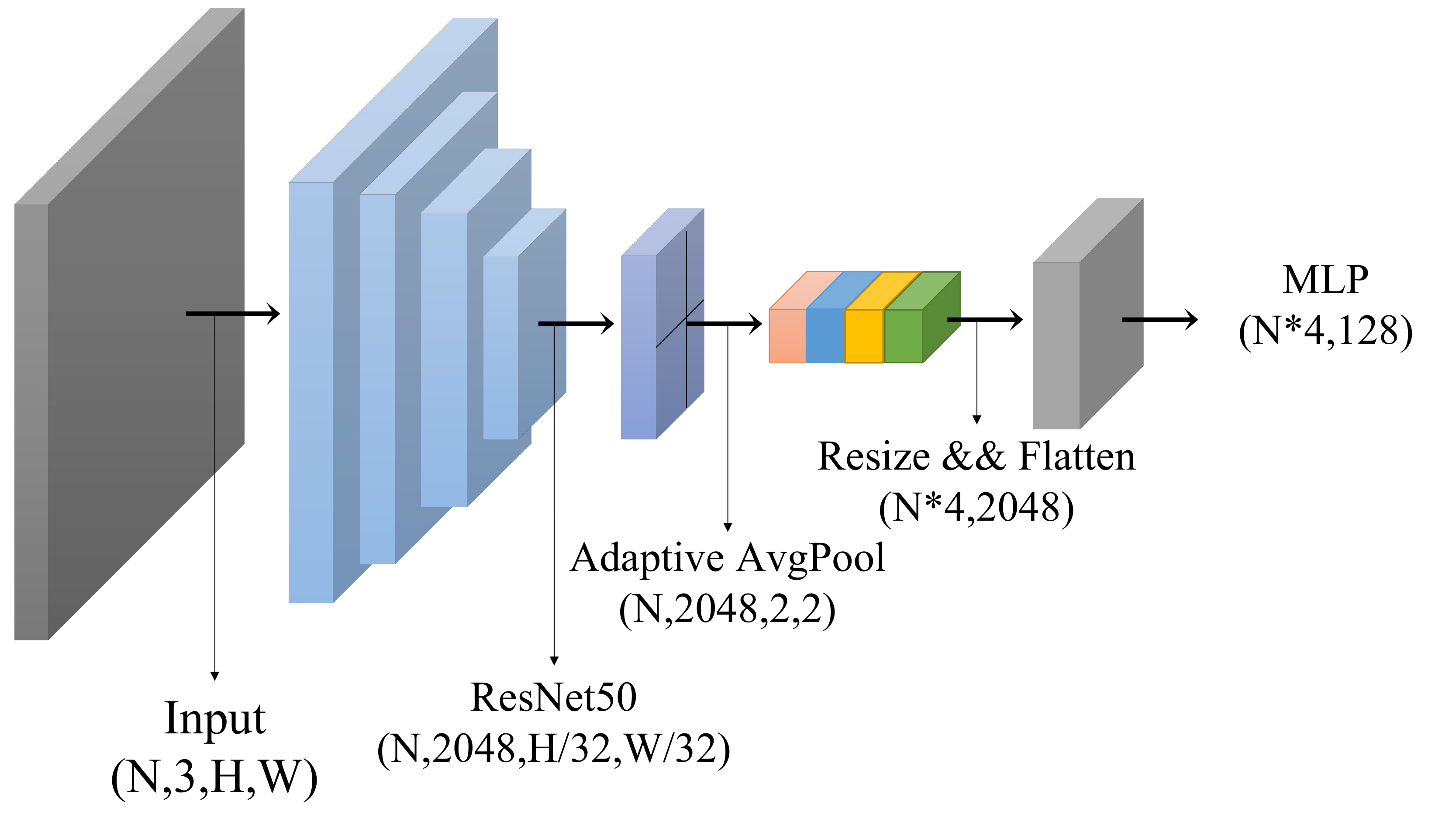}
    \caption{Illustration of multi-instance learning head module. The output of ResNet50 backbone will send into adaptive average pooling layer and generate multi-instance representations. With non-linear MLP layers, each representation will become a feature vector with 128 channels.}
    \label{fig:multi-instance}
\end{figure}
This section mainly introduces the extra learning heads named multi-instance learning head and cell-instance learning head, which we propose based on MoCo v2. The multi-instance learning head introduces multiple instances learning para-digm, which aims to boost performance on object recognition and localization. The cell-instance learning head is proposed to focus on fine-grained features to precisely locate object boundaries and extract object contents. 

\noindent\textbf{Multi-instance learning head.} Instance discrimination task in MoCo v2\cite{chen2020improved} has proved its capability in the classification task. However, it is limited to single instance learning scenarios. In this paper, we call it single-instance learning head, which can not fully mine the potential information in the image. Therefore, we propose a new multi-instance learning head in parallel with the single-instance learning head. In order to construct a multi-instance view, we randomly select multiple augmented views in one mini-batch to form a multi-instance view as the extra input of the query encoder. To balance training time and accuracy, a multi-instance view consists of four views. These four views are arranged in the order of upper left, upper right, lower left, and lower right. A concrete view can be seen from $q_{2}$ in Figure \ref{fig:alg structure}. The corresponding views of the above multi-instance view will be encoded by the key encoder so as to create positive keys. Identically, negative keys come from momentum-updated queue, which is consistent with MoCo v2.

In the technical implementation, we adopt an adaptive average pooling layer to generate multiple feature representations on multi-instance views, aiming at enabling the network to have the ability to focus on multiple instance feature representations and learn the instance localization information at the same time. As shown in Figure \ref{fig:multi-instance}, the visual representation coming from the last convolutional layer of ResNet50 will pass an adaptive average pooling layer to generate different feature vectors, each feature vector represents the visual representation of a certain instance which is shown with different colors in Figure \ref{fig:multi-instance}. With the resize and flatten operations, eventually, we will get multiple positive pairs and corresponding negative pairs. The loss function in this part is designed as follows:
\begin{equation}
\begin{aligned}
    &\mathcal{L}_{multi-ins}= 
    -\log \frac{\sum_{i}^{4}\exp \left(q_{m} \cdot k_{m_{i}+} / \tau\right)}{\sum_{i}^{4}\exp \left(q_{m} \cdot k_{m_{i}+}\right)+\sum_{k_{-}} \exp \left(q_{m} \cdot k_{-} / \tau\right)}
\end{aligned}
\end{equation}
Where $q_{m}$ represents the multi-instance feature representation encoded by query encoder and a visualized sample can been seen from Figure \ref{fig:alg structure} $q_{2}$. $k_{m_{i}+}$ represents the corresponding feature representation encoded by key encoder and visualized samples can been seen from Figure \ref{fig:alg structure} $k_{1,2,3,4}$. $k_{-}$ represents the feature encoded by past key encoders since the parameters in key encoders are updated by query encoder. This loss function will force multi-instance query features to match with corresponding key features while mismatching with the features stored in the momentum-updated queue.

\noindent\textbf{Cell-instance learning head.} Downstream tasks like object detection and instance segmentation not only require the model to have the capability of concentrating on multiple instances, but require the model to focus on fine-grained feature learning to precisely locate object boundaries and extract object contents. For instance, instance segmentation requires the model to classify pixel-level features. In order to better transfer self-supervised models to downstream prediction tasks, we propose an efficient cell-instance learning head. Compared with the single-instance head and multi-instance head, views in the cell-instance head can be defined as partial areas of original images. However, as the indispensable augmentation operator in self-supervised learning, random crop and resize can easily cause different views with no overlap areas, which will cause no matching cell feature between different views. To handle this problem, we add restrictions to the process of random crop. Compared with original random crop, which only returns the cropped image, we also return coordinates of cropped images and add a function to calculate the overlap area between views to judge whether the view generation is valid. If not, random crop stage will repeat itself until generating a valid overlap area. With the help of Precise ROI Pooling\cite{jiang2018acquisition} which can output specified size feature, we eventually generate strict feature matching area from overlap area with the empirically size of 3x3 in Res5 stage of ResNet50, each corresponding cell will be flattened to 1-D tensor and forced to be similar in embedding space. The visual expression can be seen from the cell-instance head module in Figure \ref{fig:alg structure}. Furthermore, we build another momentum-updated queue to prepare negative fine-grained feature samples, since the original momentum-updated queue used in MoCo v2 can not apply to cell-instance learning. The overall cell-instance learning head is integrated with the single-instance learning head and multi-instance learning head in parallel.

Particularly, MLP layers can only work on 3-D tensors while the fine-grained features have 4 dimensions. Therefore, we use successive 1x1 convolution layer, RELU, and 1x1 convolutional layer to replace the original MLP layers. To better represent the fine-grained feature, the final representation has 256 channels compared with MoCo v2 which output visual representation with 128 channels. Considering the differences between the cell-instance head and the other two heads, we additionally create a new momentum-updated queue for storing fine-grained features encoded by key encoder. The loss function in cell-instance learning head is defined as follows:
\begin{small}
\begin{equation}
\begin{aligned}
    &\mathcal{L}_{cell-ins}= 
    -\log\!\frac{\sum_{i}^{9}\!\exp\left(q_{c_{i}}\cdot k_{c_{i}+} / \tau\right)}{\sum_{i}^{9}\!\exp\left(q_{c_{i}}\cdot k_{c_{i}+}\right)+\sum_{i}^{9}\!\sum_{k_{c-}}\!\exp\left(q_{c_{i}}\cdot k_{c-} / \tau\right)}
\end{aligned}
\end{equation}
\end{small}
Where the $q_{c_{i}}$ represents the one cell feature representation encoded by query encoder, $k_{c_{i}+}$ represents the corresponding cell feature representation encoded by key encoder. $k_{c-}$ represents the feature stored in the fine-grained momentum-updated queue. Each overlap view will generate nine cells and each cell maps to a certain area in the original input. In this loss, the target is shorten the distance between positive cells and push the negative cells far away.
\subsection{Loss function.}
\label{subsec:loss func}
We combine the loss functions of the above three heads. The loss of each part independently adds up to the total loss with the same proportion, total loss is as follows:
\begin{small}
\begin{equation}
\begin{aligned}
&\mathcal{L}_{total}= \mathcal{L}_{moco} + \mathcal{L}_{multi-ins} + \mathcal{L}_{cell-ins} \\
&=-\log \frac{\exp \left(q \cdot k_{+} / \tau\right)}{\exp \left(q \cdot k_{+}\right)+\sum_{k_{-}} \exp \left(q \cdot k_{-} / \tau\right)} \\
&-\log \frac{\sum_{i}^{4}\exp \left(q_{m} \cdot k_{m_{i}+} / \tau\right)}{\sum_{i}^{4}\exp \left(q_{m} \cdot k_{m_{i}+}\right)+\sum_{k_{-}} \exp \left(q_{m} \cdot k_{-} / \tau\right)} \\
&-\log\!\frac{\sum_{i}^{9}\!\exp\left(q_{c_{i}}\cdot k_{c_{i}+} / \tau\right)}{\sum_{i}^{9}\!\exp\left(q_{c_{i}}\cdot k_{c_{i}+}\right)+\sum_{i}^{9}\!\sum_{k_{c-}}\!\exp\left(q_{c_{i}}\cdot k_{c-} / \tau\right)}
\end{aligned}
\end{equation}
\end{small}
In addition, we also tried to assign different weights to the total loss for each learning head. The performance did not increase but decreased. From the view of numerical value analysis, the initial loss of each part is similar and the loss reduction trend is similar as well. Therefore, the loss setting with a same radio of each learning head is reasonable.
\section{Experiments}
\label{sec:exp}
In this section, we will report the results of our proposed method on several widely used benchmarks. We transfer the self-supervised trained model to fine-grained downstream tasks including object detection and instance segmentation. Moreover, we evaluate the classification under the same settings as MoCo v2. In order to quantify the experimental results, we also conduct a series of ablation experiments.
\subsection{Experimental Settings}
\label{subsec: exp setting}
\noindent\textbf{Datasets.} All experiments involves three datasets, namely ImageNet-1K\cite{deng2009imagenet}, COCO\cite{lin2015microsoft}, PASCAL VOC\cite{everingham2010pascal}. ImageNet-1K contains \textasciitilde1.28 million image data of total 1000 classes. It was widely used to evaluate model performance on the image classification task. COCO is a large-scale object detection, segmentation and captioning dataset, which contains 328k images with 2500k labeled objects. In this paper, we use COCO2017 which contains \textasciitilde118k training images and 5k validation images. PASCAL VOC is a widely used dataset for evaluating object detection as well. Compared with ImageNet-1K, the latter two contain more challenging scenarios. For instance, in COCO, the average number of objects is about 7.6 per image, while this value is only 1.1 in ImageNet.

\noindent\textbf{Pre-training configurations.} To keep consistency with MoCo v2, we adopt the SGD optimizer with 0.0001 weight decay and 0.9 momentum. All experiments are conducted on 8 Tesla V100 GPUs with mini-batch size of 256 for 200 epochs pre-training. The initial learning rate is set to 0.03 with cosine learning rate decay\cite{loshchilov2016sgdr}. In multi-instance head and cell-instance learning head, we still use 0.2 as the temperature coefficient. The size of momentum-updated queue in the cell-instance head is set to 64512 compared with the 65536 set in the single-instance head and multi-instance head. Specifically, all crucial parameters' settings can be found in table \ref{tab:params}.

\vspace{-0.5cm}
\begin{table}[h]
\footnotesize
\caption{\textbf{Parameter Settings.} This table presents all important parameters we used for the self-supervised pre-training stage, which also makes a detailed comparison with MoCo v2. The main differences are in learning head settings. In addition, we add a new momentum-updated queue for cell-instance learning. }
\centering
\setlength{\tabcolsep}{6mm}{
\renewcommand\arraystretch{1.0}
\begin{tabular}{lll}
\hline
\textbf{Parameters}                              & \textbf{MoCo v2} & \textbf{Ours}                               \\ \hline
batch size                                       & 256              & 256                                         \\ 
GPU                                              & 8 * Tesla V100   & 8 * Tesla V100                              \\ 
optimizer                                        & SGD              & SGD                                         \\ 
learning rate                                    & 0.03             & 0.03                                        \\ 
weight decay                                     & 0.0001           & 0.0001                                      \\ 
momentum                                         & 0.9              & 0.9                                         \\ 
epoch                                            & 200              & 200                                         \\ 
temperature                                      & 0.2              & 0.2                                         \\ 
\multirow{1}{*}{feature dimension}               & single-ins: 128   & single-ins: 128                              \\  
                                                 & -                & multi-ins: 128                    \\  
                                                 & -                & cell-ins: 256              \\ 
\multirow{1}{*}{k}                               & 65536            & queue1: 65536              \\ 
                                                 & -                & queue2: 64512 \\ \hline
\end{tabular}}
\label{tab:params}
\end{table}

\vspace{-0.8cm}
\subsection{Main Results. }
\label{subsec: main res}

\noindent\textbf{Evaluation protocols.} We evaluate the performance by transferring weights to downstream prediction tasks. To align the results with other methods, we conduct a series of experiments on widely used datasets including PASCAL VOC object detection, COCO object detection, COCO instance segmentation. Specifically, ResNet50 with a fully connected layer is adopted to evaluate the performance on ImageNet classification task. Mask R-CNN(ResNet50+FPN\cite{lin2017feature} backbone) is adopted to evaluate the performance on COCO object detection and instance segmentation with standard 1x schedule and 2x schedule. We also adopt Mask R-CNN(ResNet50+C4 backbone) to evaluate the performance on COCO object detection and instance segmentation with standard 1x schedule. Faster R-CNN(ResNet50+C4 backbone) is adopted to evaluate the performance on PASCAL VOC object detection. Network structures and related configurations for downstream tasks come from Detectron2 with an initial learning rate 0.02 and batch size of 16. Average precision on bounding boxes named AP$^{bb}$ and average precision on masks named AP$^{mk}$ are used as main evaluation metrics. AP$_{50}$(IoU threshold is 50\%)and AP$_{75}$(IoU threshold is 75\%) are also used.

\noindent\textbf{PASCAL VOC object detection.} We strictly follow the settings used in MoCo v2, which uses Faster R-CNN ResNet50-C4 for fine-tuning. We fine-tune all layers of ResNet50 pre-trained on ImageNet. The short-side length of input images ranges from 480 to 800 pixels at the training stage and 800 at the inference stage. We report the results of fine-tuning on PASCAL VOC trainval07+12 and evaluating on PASCAL VOC test2007 in table \ref{tab:voc detec}. Compared with previous state-of-the-arts, specifically, InsCon surpasses its baseline MoCo v2 by 2.1\%/1.2\%/3.0\% at AP$^{bb}$/AP$^{bb}_{50}$/AP$^{bb}_{75}$.

\vspace{-0.5cm}
\begin{table}[ht]
\footnotesize
\caption{\textbf{Object detection results on PASCAL VOC datasets with Faster R-CNN R50-C4 for 24k iterations.} Fine-tuning stage is conducted on VOC07+12 trainval data and evaluating on VOC07 test data. InsCon achieves state-of-the-art results.}
\centering
\setlength{\tabcolsep}{7mm}{
\renewcommand\arraystretch{0.9}
\begin{tabular}{c|c|ccc}
\multirow{2}{*}{\textbf{Method}} & \multirow{2}{*}{\textbf{Epoch}} & \multicolumn{3}{c}{\textbf{Faster R-CNN R50-C4}}                    \\ \cline{3-5} 
                                 &                                 & \multicolumn{1}{c}{AP$^{bb}$}         & \multicolumn{1}{c}{AP$^{bb}_{50}$}       & AP$^{bb}_{75}$       \\ \hline
random init                      & -                               & \multicolumn{1}{c}{32.8}       & \multicolumn{1}{c}{59.0}       & 31.6       \\ 
super. IN                        & -                               & \multicolumn{1}{c}{54.2}       & \multicolumn{1}{c}{81.6}       & 59.8       \\ \hline
MoCo                             & 200                             & \multicolumn{1}{c}{55.9} & \multicolumn{1}{c}{81.5} & 62.6 \\ 
MoCo v2                          & 200                             & \multicolumn{1}{c}{57.0} & \multicolumn{1}{c}{82.4} & 63.6 \\ 
InfoMin                          & 200                             & \multicolumn{1}{c}{57.6} & \multicolumn{1}{c}{82.7} & 64.6 \\ 
DetCo                            & 200                             & \multicolumn{1}{c}{57.8} & \multicolumn{1}{c}{82.6} & 64.2 \\ 
ReSim                            & 200                             & \multicolumn{1}{c}{58.7} & \multicolumn{1}{c}{83.1} & 66.3 \\ 
DenseCL                          & 200                             & \multicolumn{1}{c}{58.7} & \multicolumn{1}{c}{82.8} & 65.2 \\ \hline
InsCon                             & 200                             & \multicolumn{1}{c}{\textbf{59.1}}           & \multicolumn{1}{c}{\textbf{83.6}}           &  \textbf{66.6}         \\ 
\end{tabular}}
\label{tab:voc detec}
\end{table}

\vspace{-1.0cm}
\begin{table*}[h]
\footnotesize
\caption{\textbf{Object detection and instance segmentation results on COCO with Mask R-CNN R50-C4 for 90k iterations.} All models are pre-trained 200 epochs on ImageNet. Fine-tuning stage on COCO2017 is reimplemented in same configurations for 90k iterations. InsCon outperforms both supervised method and self-supervised methods.}
\centering
\setlength{\tabcolsep}{2.8mm}{
\renewcommand\arraystretch{1.0}
\begin{tabular}{c|c|cccccc}
\multirow{2}{*}{\textbf{Method}} & \multirow{2}{*}{\textbf{Epoch}} & \multicolumn{6}{c}{\textbf{Mask R-CNN R50-C4 1x schedule}}                                                                                                        \\ \cline{3-8} 
                                 &                                 & \multicolumn{1}{c}{AP$^{bb}$}   & \multicolumn{1}{c}{AP$^{bb}_{50}$} & \multicolumn{1}{c|}{AP$^{bb}_{75}$} & \multicolumn{1}{c}{AP$^{mk}$}   & \multicolumn{1}{c}{AP$^{mk}_{50}$} & AP$^{mk}_{75}$                  \\ \hline
Supervised                            & -                             & \multicolumn{1}{c}{38.2} & \multicolumn{1}{c}{58.2} & \multicolumn{1}{c|}{41.2} & \multicolumn{1}{c}{33.3} & \multicolumn{1}{c}{54.7} & 35.2                  \\ \hline
MoCo\cite{he2020momentum}                             & 200                             & \multicolumn{1}{c}{38.4} & \multicolumn{1}{c}{58.2} & \multicolumn{1}{c|}{41.7} & \multicolumn{1}{c}{33.6} & \multicolumn{1}{c}{54.9} & 35.7                 \\
MoCo v2\cite{chen2020improved}                          & 200                             & \multicolumn{1}{c}{38.9} & \multicolumn{1}{c}{58.7} & \multicolumn{1}{c|}{41.9} & \multicolumn{1}{c}{34.2} & \multicolumn{1}{c}{55.4} & 36.2                 \\
ReSim\cite{xiao2021region}                            & 200                             & \multicolumn{1}{c}{39.7} & \multicolumn{1}{c}{59.0} & \multicolumn{1}{c|}{43.0} & \multicolumn{1}{c}{34.6} & \multicolumn{1}{c}{55.9} & 37.1                 \\
DetCo\cite{xie2021detco}                            & 200                             & \multicolumn{1}{c}{39.8} & \multicolumn{1}{c}{59.7} & \multicolumn{1}{c|}{43.0} & \multicolumn{1}{c}{34.7} & \multicolumn{1}{c}{56.3} & 36.7                  \\ \hline
InsCon                             &200                                 & \multicolumn{1}{c}{\textbf{40.3}}     & \multicolumn{1}{c}{\textbf{60.0}}     & \multicolumn{1}{c|}{\textbf{43.5}}     & \multicolumn{1}{c}{\textbf{35.1}}     & \multicolumn{1}{c}{\textbf{56.7}}     & \multicolumn{1}{c}{\textbf{37.6}} \\
\end{tabular}}
\label{tab:coco res c4}
\end{table*}

\noindent\textbf{COCO object detection and instance segmentation. }
We adopt the ResNet-50-C4 and Mask R-CNN ResNet50-FPN to evaluate results on COCO2017. Respectively, all layers in ResNet50 transferred in fine-tuning stage and all results are evaluated in same configurations. The short-side length of images ranges from 640 to 800 pixels during the training stage and fixes at 800 during the inference stage. We report the results that fine-tune on train2017 with \textasciitilde118k images and evaluate performance on val2017 as shown in Table \ref{tab:coco res c4}, Table \ref{tab:coco res fpn 1x}, Table \ref{tab:coco res fpn 2x}. Compared with previous work, respectively, InsCon with Mask R-CNN R50-FPN and 1x training schedule gets object detection results with 40.9\% at AP$^{bb}$, which surpasses MoCo v2 by 1.1\%. Furthermore, InsCon gets results on instance segmentation task with 36.9\% at AP$^{mk}$, which surpasses MoCo v2 by 1.0\%. Specifically, InsCon gets 42.8\% at AP$^{bb}$ and 38.5\% at AP$^{mk}$ with Mask R-CNN R50-FPN and 2x training schedule. All these results achieve state-of-the-art performance.

\begin{figure}[h]
    \centering
    \includegraphics[width=1.0\linewidth]{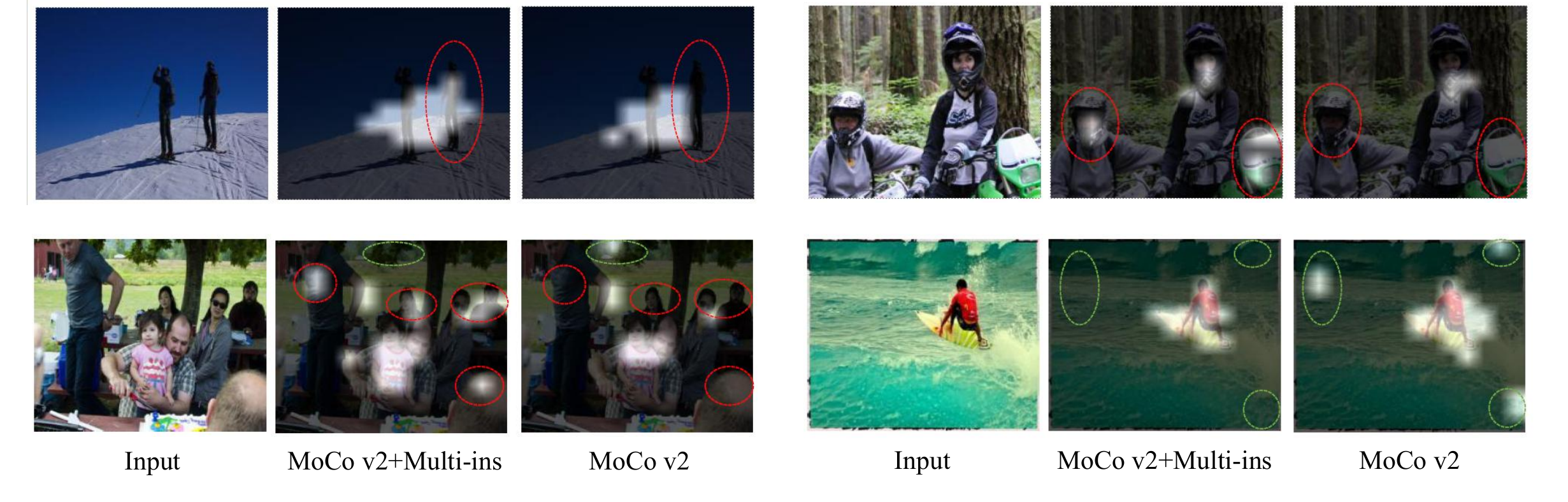}
    \caption{\textbf{Attention map comparison between MoCo v2 and MoCo v2 with multi-instance learning head.} Red circles explain that our method can recognize and locate more objects. Green circles demonstrate that our method can precisely locate objects.}
    \label{fig:vis}
\end{figure}

\noindent\textbf{Linear classification. }
To demonstrate the ability of multi-instance learning head in capturing instances in complex scenarios, we verify our method by linear classification on frozen features, following the protocol used in MoCo v2, we freeze ResNet50 parameters trained by 200 epochs self-supervised contrastive learning and train a linear classifier(a linear layer followed by softmax ) with 100 epochs. In addition, the initial learning rate is 30 which is consistent with MoCo v2. We report the results in Table \ref{tab:cls}. Compared with baseline MoCo v2, we get 2.8\% improvement in top-1 accuracy. which also surpasses SimCLR v1 by 3.7\% top-1 accuracy. Moreover, we randomly select some images from COCO2017 validation images and visualize the attention map generated by ResNet50 with different weights. Figure \ref{fig:vis} demonstrates the effectiveness of our multi-instance learning head.

\vspace{-0.5cm}
\begin{table*}[ht]
\footnotesize
\caption{\textbf{Object detection and instance segmentation results on COCO with Mask R-CNN R50-FPN for 90k iterations.} All models except PixPro are pre-trained 200 epochs on ImageNet. PixPro is pre-trained 400 epochs on ImageNet which obtained by official model. Fine-tuning stage on COCO2017 is reimplemented in same configurations for 90k iterations. InsCon outperforms both supervised method and self-supervised methods. The results are averaged over 5 independent trials.}
\centering
\setlength{\tabcolsep}{2.8mm}{
\renewcommand\arraystretch{1.0}
\begin{tabular}{c|c|cccccc}
\multirow{2}{*}{\textbf{Method}} & \multirow{2}{*}{\textbf{Epoch}} & \multicolumn{6}{c}{\textbf{Mask R-CNN R50-FPN 1x schedule}}                                                                                                        \\ \cline{3-8} 
                                 &                                 & \multicolumn{1}{c}{ AP$^{bb}$}   & \multicolumn{1}{c}{AP$^{{bb}}_{50}$} & \multicolumn{1}{c|}{AP$^{bb}_{75}$} & \multicolumn{1}{c}{AP$^{mk}$}   & \multicolumn{1}{c}{AP$^{mk}_{50}$} & AP$^{mk}_{75}$                  \\ \hline
Supervised                            & -                             & \multicolumn{1}{c}{39.7} & \multicolumn{1}{c}{59.5} & \multicolumn{1}{c|}{43.3} & \multicolumn{1}{c}{35.9} & \multicolumn{1}{c}{56.6} & 38.6                  \\ \hline
MoCo\cite{he2020momentum}                             & 200                             & \multicolumn{1}{c}{39.3} & \multicolumn{1}{c}{59.0} & \multicolumn{1}{c|}{43.0} & \multicolumn{1}{c}{35.6} & \multicolumn{1}{c}{56.0} & 38.2                  \\
MoCo v2\cite{chen2020improved}                          & 200                             & \multicolumn{1}{c}{39.8} & \multicolumn{1}{c}{59.7} & \multicolumn{1}{c|}{43.7} & \multicolumn{1}{c}{35.9} & \multicolumn{1}{c}{56.7} & 38.6                  \\
ReSim\cite{xiao2021region}                            & 200                             & \multicolumn{1}{c}{40.0} & \multicolumn{1}{c}{59.6} & \multicolumn{1}{c|}{43.9} & \multicolumn{1}{c}{36.2} & \multicolumn{1}{c}{56.7} & 38.8                 \\
DenseCL\cite{wang2021dense}                          & 200                             & \multicolumn{1}{c}{40.1} & \multicolumn{1}{c}{59.9} & \multicolumn{1}{c|}{43.6} & \multicolumn{1}{c}{36.2} & \multicolumn{1}{c}{56.9} & 39.2                  \\
DetCo\cite{xie2021detco}                            & 200                             & \multicolumn{1}{c}{40.1} & \multicolumn{1}{c}{60.1} & \multicolumn{1}{c|}{44.0} & \multicolumn{1}{c}{36.4} & \multicolumn{1}{c}{57.4} & 39.1                  \\
PixPro\cite{xie2021propagate}                            & 400                             & \multicolumn{1}{c}{40.6} & \multicolumn{1}{c}{60.3} & \multicolumn{1}{c|}{44.6} & \multicolumn{1}{c}{36.8} & \multicolumn{1}{c}{57.5} & 39.6                  \\ \hline
InsCon                             &200                                 & \multicolumn{1}{c}{\textbf{40.9}}     & \multicolumn{1}{c}{\textbf{60.9}}     & \multicolumn{1}{c|}{\textbf{44.8}}     & \multicolumn{1}{c}{\textbf{36.9}}     & \multicolumn{1}{c}{\textbf{57.9}}     & \multicolumn{1}{c}{\textbf{39.8}} \\
\end{tabular}}
\label{tab:coco res fpn 1x}
\end{table*}

\vspace{-0.5cm}
\begin{table*}[h]
\footnotesize
\caption{\textbf{Object detection and instance segmentation results on COCO with Mask R-CNN R50 FPN for 180k iterations.} All models are pre-trained 200 epochs on ImageNet. Fine-tuning stage on COCO2017 is reimplemented in same configurations for 180k iterations. InsCon outperforms both supervised method and self-supervised methods.}
\centering
\setlength{\tabcolsep}{2.8mm}{
\renewcommand\arraystretch{1.0}
\begin{tabular}{c|c|cccccc}
\multirow{2}{*}{\textbf{Method}} & \multirow{2}{*}{\textbf{Epoch}} & \multicolumn{6}{c}{\textbf{Mask R-CNN R50-FPN 2x schedule}}                                                                                                        \\ \cline{3-8} 
                                 &                                 & \multicolumn{1}{c}{ AP$^{bb}$}   & 
                                 \multicolumn{1}{c}{AP$^{bb}_{ 50}$} & \multicolumn{1}{c|}{AP$^{bb}_{75}$} & \multicolumn{1}{c}{AP$^{mk}$}   & \multicolumn{1}{c}{AP$^{mk}_{50}$} & AP$^{mk}_{75}$                  \\ \hline
Supervised                            & -                             & \multicolumn{1}{c}{42.0} & \multicolumn{1}{c}{61.9} & \multicolumn{1}{c|}{46.0} & \multicolumn{1}{c}{38.0} & \multicolumn{1}{c}{59.2} & 41.1                  \\ \hline
MoCo\cite{he2020momentum}                             & 200                             & \multicolumn{1}{c}{41.5} & \multicolumn{1}{c}{61.3} & \multicolumn{1}{c|}{45.4} & \multicolumn{1}{c}{37.4} & \multicolumn{1}{c}{58.6} & 40.2                  \\
MoCo v2\cite{chen2020improved}                          & 200                             & \multicolumn{1}{c}{41.9} & \multicolumn{1}{c}{61.9} & \multicolumn{1}{c|}{45.8} & \multicolumn{1}{c}{37.8} & \multicolumn{1}{c}{59.2} & 40.9                  \\
ReSim\cite{xiao2021region}                            & 200                             & \multicolumn{1}{c}{42.3} & \multicolumn{1}{c}{62.2} & \multicolumn{1}{c|}{46.4} & \multicolumn{1}{c}{38.2} & \multicolumn{1}{c}{59.6} & 41.0                 \\
DenseCL\cite{wang2021dense}                          & 200                             & \multicolumn{1}{c}{42.2} & \multicolumn{1}{c}{62.4} & \multicolumn{1}{c|}{46.1} & \multicolumn{1}{c}{38.1} & \multicolumn{1}{c}{59.4} & 40.9                  \\
DetCo\cite{xie2021detco}                            & 200                             & \multicolumn{1}{c}{42.3} & \multicolumn{1}{c}{62.4} & \multicolumn{1}{c|}{46.0} & \multicolumn{1}{c}{38.0} & \multicolumn{1}{c}{59.4} & 40.9                  \\ \hline
InsCon                             &200                                 & \multicolumn{1}{c}{\textbf{42.8}}     & \multicolumn{1}{c}{\textbf{62.8}}     & \multicolumn{1}{c|}{\textbf{46.5}}     & \multicolumn{1}{c}{\textbf{38.5}}     & \multicolumn{1}{c}{\textbf{60.5}}     & \multicolumn{1}{c}{\textbf{41.4}} \\
\end{tabular}}
\label{tab:coco res fpn 2x}
\end{table*}

\vspace{-0.5cm}
\subsection{Ablation study. }
\label{subsec: abla}
In this subsection, we conduct a series of extensive ablation studies to represent how each component contributes to InsCon. Besides, we also explore the function of hyper-parameter settings and the influence of different cell-instance learning positions. 

\noindent\textbf{Comparison of multi-instance and cell-instance head. }
In order to quantify the function of each learning head. Based on MoCo v2, we separately add each module to the model to record the results. Multi-instance learning head achieves significant improvement not only on ImageNet classification as shown in Table \ref{tab:cls}, but also on object detection and instance segmentation as shown in Table \ref{tab:module}. Cell-instance learning head improves object detection and instance segmentation performances by 0.5\% AP$^{bb}$ and 0.5\% AP$^{mk}$ as shown in Table \ref{tab:module}. When combining all modules, performances on downstream prediction tasks are further boosted.

\noindent\textbf{Temperature coefficient for cell-instance learning head.} The temperature coefficient controls the sensitivity of contrastive loss in self-supervised training, which can be explained by gradient values in \cite{wang2021understanding}. Some related calculating formulas are as follows:
\begin{equation}
\frac{\partial \mathcal{L}\left(x_{i}\right)}{\partial s_{i, i}}=-\frac{1}{\tau} \sum_{k \neq i} P_{i, k}, \quad \frac{\partial \mathcal{L}\left(x_{i}\right)}{\partial s_{i, j}}=\frac{1}{\tau} P_{i, j}
\end{equation}
\begin{equation}
P_{i, j}=\frac{\exp \left(s_{i, j} / \tau\right)}{\sum_{k \neq i} \exp \left(s_{i, k} / \tau\right)+\exp \left(s_{i, i} / \tau\right)}
\end{equation}
Where $s_{i,i}$ represents the similarity of positive pairs, $s_{i,j}$ represents the similarity of negative pairs. $P_{i,j}$ represents the contrastive loss formula. From the derivation results of positive pairs and negative pairs, we can adjust temperature coefficient to change the attention ratio on positive pairs and negative pairs. To quantify this procedure, Table \ref{tab:temp} examines the sensitivity of the temperature coefficient used in the cell-instance learning head. We test the temperature coefficient from 0.1 to 0.4, and we find the peak value 
when {$\tau_{cell-ins}$} equals 0.2. In this value, AP$^{bb}$ is greater than the upper adjacent value by 0.4\%, AP$^{bb}_{50}$ is greater than the upper and lower adjacent values by 0.5\% and 0.3\%, and AP$^{bb}_{75}$ is greater than the upper and lower adjacent values by 0.5\% and 0.2\%.

\vspace{-0.5cm}
\begin{table}[h]
\footnotesize
\caption{\textbf{Top-1 accuracy of linear clssification on ImageNet with multi-instance learning head.} `multi-ins' represents the multi-instance learning head. All models are pre-trained 200 epochs on ImageNet and fine-tuned 100 epoches with a linear classifier.}
\centering
\setlength{\tabcolsep}{10mm}{
\renewcommand\arraystretch{1.0}
\begin{tabular}{cc}
\hline
\textbf{Method}          & \textbf{Classification(\%)} \\ \hline
MoCo\cite{he2020momentum}                     & 60.6                        \\ 
SeLa\cite{asano2019self}                     & 61.5                        \\ 
PIRL\cite{misra2020self}                     & 63.6                        \\ 
CPC v2\cite{henaff2020data}                   & 63.8                        \\ 
PCL\cite{li2020prototypical}                      & 65.9                        \\ 
SimCLR v1\cite{chen2020simple}                & 66.6                        \\ 
MoCo v2\cite{chen2020improved}                  & 67.5                        \\ \hline
MoCo v2 + multi-ins & \textbf{70.3}               \\ \hline
\end{tabular}}
\label{tab:cls}
\end{table}

\vspace{-1.5cm}
\begin{table}[h]
\footnotesize
\caption{\textbf{Ablation study of different learning heads on COCO with Mask R-CNN R50-FPN for 90k iterations.} `multi-ins head' represents the multi-instance learning head, `cell-ins head' represents the cell-instance learning head.}
\centering
\setlength{\tabcolsep}{5mm}{
\renewcommand\arraystretch{0.9}
\begin{tabular}{c|ccccc}
\textbf{Method} & {\begin{tabular}[c]{@{}c@{}}multi-ins head\end{tabular}} & {\begin{tabular}[c]{@{}c@{}}cell-ins head\end{tabular}} & {AP$^{bb}$} & {AP$^{mk}$} \\ \hline
MoCo v2\cite{chen2020improved}         &                                                                         &                                                                               & 39.8        & 35.9                   \\ 
Ours            &                                                                       & \checkmark                                                                              & 40.3 & 36.4      \\ 
Ours            & \checkmark                                                                         &                                                                       & 40.7 & 36.6     \\ 
Ours            & \checkmark                                                                      & \checkmark                                                                          &\textbf{40.9}             &\textbf{36.9}                              \\ \hline
\end{tabular}}
\label{tab:module}
\end{table}

\begin{table}[h]
\footnotesize
\caption{\textbf{Ablation study of different temperature coefficient on COCO object detection with Mask R-CNN FPN for 90k iterations.} $\tau_{cell-ins}$ represents the temperature coefficient used in cell-instance learning head, and $\tau = 0.2$ or $0.3$ shows the best performance in object detection.}
\centering
\setlength{\tabcolsep}{7mm}{
\renewcommand\arraystretch{0.9}
\begin{tabular}{clcccc}
\multicolumn{2}{c|}{\textbf{Method}}                                                                            & {$\tau_{cell-ins}$} & {AP$^{bb}$} & {AP$^{bb}_{50}$} & {AP$^{bb}_{75}$} \\ \hline
\multicolumn{2}{c|}{\multirow{4}{*}{\begin{tabular}[c]{@{}c@{}}MoCo v2 \\ +\\ cell-ins head\end{tabular}}} & 0.1        & 39.9        & 59.6          & 43.7          \\ 
\multicolumn{2}{c|}{}                                                                                           & 0.2        & \textbf{40.3}        & \textbf{60.1}          & \textbf{44.2}          \\ 
\multicolumn{2}{c|}{}                                                                                           & 0.3        & \textbf{40.3}        & 59.8          & 44.0          \\ 
\multicolumn{2}{c|}{}                                                                                           & 0.4        & 40.2        & 59.7          & 44.0          \\ \hline
\end{tabular}}
\label{tab:temp}
\end{table}

\begin{table}[h]
\footnotesize
\caption{\textbf{Ablation study of cell-instance learning head positions.} `Res4' represents that cell-instance learning head is added at Res4 stage of ResNet50. `Res5' represents that cell-instance learning head is added at Res5 stage.}
\centering
\setlength{\tabcolsep}{7mm}{
\renewcommand\arraystretch{0.9}
\begin{tabular}{c|clll}
\textbf{Method} & {cell-ins  position} & \multicolumn{1}{c}{{AP$^{bb}$}} & \multicolumn{1}{c}{{AP$^{mk}$}} &  \\ \hline
Ours            &Res4                    & 40.5                                 & 36.6                                                                       \\ 
Ours            &Res5                  & \textbf{40.9}                                 & \textbf{36.9}                                                                       \\ \hline
\end{tabular}}
\label{tab:position}
\end{table}

\noindent\textbf{Positions for cell-instance learning head.}
Each stage in ResNet scales the output features of the previous stage. In ResNet50, the scale ratio is 2x, 4x, 8x, 16x and 32x. Thus, cell features generated by each stage map to different size regions in original images. For instance, one cell in res5 represents 32x32 regions and in res4 represents 16x16 regions, which represents different semantic features. It means the views generated in cell-instance learning head are also affected by cell feature generating positions. As shown in Table \ref{tab:position}, we explore the influence of positions in cell-instance learning head for final results. When cell-instance learning head is connected after res5, AP$^{bb}$ is 0.4\% greater than connecting after the res4, and AP$^{mk}$ is 0.3\% greater than connecting after the res4.

\section{Conclusion}
\label{sec:conclude}
In this paper, we present InsCon, a general and effective end-to-end learning framework for feature representation from large-scale unlabeled images. InsCon is committed to unleashing the potential of self-supervised learning via our newly designed pretext named instance consistency. Firstly, InsCon introduces a multi-instance learning head aiming to improve the model's ability to recognize and locate multiple objects, which brings significant improvement in classification, object detection, and instance segmentation. Secondly, InsCon introduces a cell-instance learning head aiming to extract fine-grained feature for precise boundary localization, which further boost the performance on object detection and instance segmentation. Eventually, combining the above two modules with the single-instance learning head in MoCo v2, InsCon demonstrates state-of-the-art performance on various downstream prediction tasks.



%
%
\bibliographystyle{splncs04}
\bibliography{egbib}

\begin{thebibliography}{10}
\providecommand{\url}[1]{\texttt{#1}}
\providecommand{\urlprefix}{URL }
\providecommand{\doi}[1]{https://doi.org/#1}

\bibitem{asano2019self}
Asano, Y., Rupprecht, C., Vedaldi, A.: Self-labelling via simultaneous
  clustering and representation learning. In: International Conference on
  Learning Representations (2019)

\bibitem{bachman2019learning}
Bachman, P., Hjelm, R.D., Buchwalter, W.: Learning representations by
  maximizing mutual information across views. arXiv preprint arXiv:1906.00910
  (2019)

\bibitem{mmdetection}
Chen, K., Wang, J., Pang, J., Cao, Y., Xiong, Y., Li, X., Sun, S., Feng, W.,
  Liu, Z., Xu, J., Zhang, Z., Cheng, D., Zhu, C., Cheng, T., Zhao, Q., Li, B.,
  Lu, X., Zhu, R., Wu, Y., Dai, J., Wang, J., Shi, J., Ouyang, W., Loy, C.C.,
  Lin, D.: {MMDetection}: Open mmlab detection toolbox and benchmark. arXiv
  preprint arXiv:1906.07155  (2019)

\bibitem{chen2020simple}
Chen, T., Kornblith, S., Norouzi, M., Hinton, G.: A simple framework for
  contrastive learning of visual representations. In: International conference
  on machine learning. pp. 1597--1607. PMLR (2020)

\bibitem{chen2020improved}
Chen, X., Fan, H., Girshick, R., He, K.: Improved baselines with momentum
  contrastive learning. arXiv preprint arXiv:2003.04297  (2020)

\bibitem{deng2009imagenet}
Deng, J., Dong, W., Socher, R., Li, L.J., Li, K., Fei-Fei, L.: Imagenet: A
  large-scale hierarchical image database. In: 2009 IEEE conference on computer
  vision and pattern recognition. pp. 248--255. Ieee (2009)

\bibitem{dosovitskiy2020image}
Dosovitskiy, A., Beyer, L., Kolesnikov, A., Weissenborn, D., Zhai, X.,
  Unterthiner, T., Dehghani, M., Minderer, M., Heigold, G., Gelly, S., et~al.:
  An image is worth 16x16 words: Transformers for image recognition at scale.
  arXiv preprint arXiv:2010.11929  (2020)

\bibitem{everingham2010pascal}
Everingham, M., Van~Gool, L., Williams, C.K., Winn, J., Zisserman, A.: The
  pascal visual object classes (voc) challenge. International journal of
  computer vision  \textbf{88}(2),  303--338 (2010)

\bibitem{grill2020bootstrap}
Grill, J.B., Strub, F., Altch{\'e}, F., Tallec, C., Richemond, P.H.,
  Buchatskaya, E., Doersch, C., Pires, B.A., Guo, Z.D., Azar, M.G., et~al.:
  Bootstrap your own latent: A new approach to self-supervised learning. arXiv
  preprint arXiv:2006.07733  (2020)

\bibitem{hadsell2006dimensionality}
Hadsell, R., Chopra, S., LeCun, Y.: Dimensionality reduction by learning an
  invariant mapping. In: 2006 IEEE Computer Society Conference on Computer
  Vision and Pattern Recognition (CVPR'06). vol.~2, pp. 1735--1742. IEEE (2006)

\bibitem{he2020momentum}
He, K., Fan, H., Wu, Y., Xie, S., Girshick, R.: Momentum contrast for
  unsupervised visual representation learning. In: Proceedings of the IEEE/CVF
  Conference on Computer Vision and Pattern Recognition. pp. 9729--9738 (2020)

\bibitem{he2017mask}
He, K., Gkioxari, G., Doll{\'a}r, P., Girshick, R.: Mask r-cnn. In: Proceedings
  of the IEEE international conference on computer vision. pp. 2961--2969
  (2017)

\bibitem{he2016deep}
He, K., Zhang, X., Ren, S., Sun, J.: Deep residual learning for image
  recognition. In: Proceedings of the IEEE conference on computer vision and
  pattern recognition. pp. 770--778 (2016)

\bibitem{henaff2020data}
Henaff, O.: Data-efficient image recognition with contrastive predictive
  coding. In: International Conference on Machine Learning. pp. 4182--4192.
  PMLR (2020)

\bibitem{howard2013some}
Howard, A.G.: Some improvements on deep convolutional neural network based
  image classification. arXiv preprint arXiv:1312.5402  (2013)

\bibitem{jiang2018acquisition}
Jiang, B., Luo, R., Mao, J., Xiao, T., Jiang, Y.: Acquisition of localization
  confidence for accurate object detection. In: Proceedings of the European
  conference on computer vision (ECCV). pp. 784--799 (2018)

\bibitem{li2020prototypical}
Li, J., Zhou, P., Xiong, C., Hoi, S.C.: Prototypical contrastive learning of
  unsupervised representations. arXiv preprint arXiv:2005.04966  (2020)

\bibitem{lin2017feature}
Lin, T.Y., Doll{\'a}r, P., Girshick, R., He, K., Hariharan, B., Belongie, S.:
  Feature pyramid networks for object detection. In: Proceedings of the IEEE
  conference on computer vision and pattern recognition. pp. 2117--2125 (2017)

\bibitem{lin2015microsoft}
Lin, T.Y., Maire, M., Belongie, S., Bourdev, L., Girshick, R., Hays, J.,
  Perona, P., Ramanan, D., Zitnick, C.L., Dollár, P.: Microsoft coco: Common
  objects in context (2015)

\bibitem{liu2021swin}
Liu, Z., Lin, Y., Cao, Y., Hu, H., Wei, Y., Zhang, Z., Lin, S., Guo, B.: Swin
  transformer: Hierarchical vision transformer using shifted windows. In:
  Proceedings of the IEEE/CVF International Conference on Computer Vision. pp.
  10012--10022 (2021)

\bibitem{loshchilov2016sgdr}
Loshchilov, I., Hutter, F.: Sgdr: Stochastic gradient descent with warm
  restarts. arXiv preprint arXiv:1608.03983  (2016)

\bibitem{misra2020self}
Misra, I., Maaten, L.v.d.: Self-supervised learning of pretext-invariant
  representations. In: Proceedings of the IEEE/CVF Conference on Computer
  Vision and Pattern Recognition. pp. 6707--6717 (2020)

\bibitem{oord2018representation}
Oord, A.v.d., Li, Y., Vinyals, O.: Representation learning with contrastive
  predictive coding. arXiv preprint arXiv:1807.03748  (2018)

\bibitem{pathak2016context}
Pathak, D., Krahenbuhl, P., Donahue, J., Darrell, T., Efros, A.A.: Context
  encoders: Feature learning by inpainting. In: Proceedings of the IEEE
  conference on computer vision and pattern recognition. pp. 2536--2544 (2016)

\bibitem{pinheiro2020unsupervised}
Pinheiro, P.O., Almahairi, A., Benmalek, R.Y., Golemo, F., Courville, A.:
  Unsupervised learning of dense visual representations. arXiv preprint
  arXiv:2011.05499  (2020)

\bibitem{ren2016faster}
Ren, S., He, K., Girshick, R., Sun, J.: Faster r-cnn: towards real-time object
  detection with region proposal networks. IEEE transactions on pattern
  analysis and machine intelligence  \textbf{39}(6),  1137--1149 (2016)

\bibitem{tian2020makes}
Tian, Y., Sun, C., Poole, B., Krishnan, D., Schmid, C., Isola, P.: What makes
  for good views for contrastive learning? arXiv preprint arXiv:2005.10243
  (2020)

\bibitem{vaswani2017attention}
Vaswani, A., Shazeer, N., Parmar, N., Uszkoreit, J., Jones, L., Gomez, A.N.,
  Kaiser, {\L}., Polosukhin, I.: Attention is all you need. Advances in neural
  information processing systems  \textbf{30} (2017)

\bibitem{vincent2008extracting}
Vincent, P., Larochelle, H., Bengio, Y., Manzagol, P.A.: Extracting and
  composing robust features with denoising autoencoders. In: Proceedings of the
  25th international conference on Machine learning. pp. 1096--1103 (2008)

\bibitem{wang2021understanding}
Wang, F., Liu, H.: Understanding the behaviour of contrastive loss. In:
  Proceedings of the IEEE/CVF Conference on Computer Vision and Pattern
  Recognition. pp. 2495--2504 (2021)

\bibitem{wang2021dense}
Wang, X., Zhang, R., Shen, C., Kong, T., Li, L.: Dense contrastive learning for
  self-supervised visual pre-training. In: Proceedings of the IEEE/CVF
  Conference on Computer Vision and Pattern Recognition. pp. 3024--3033 (2021)

\bibitem{wu2019detectron2}
Wu, Y., Kirillov, A., Massa, F., Lo, W.Y., Girshick, R.: Detectron2.
  \url{https://github.com/facebookresearch/detectron2} (2019)

\bibitem{wu2018unsupervised}
Wu, Z., Xiong, Y., Yu, S.X., Lin, D.: Unsupervised feature learning via
  non-parametric instance discrimination. In: Proceedings of the IEEE
  conference on computer vision and pattern recognition. pp. 3733--3742 (2018)

\bibitem{xiao2021region}
Xiao, T., Reed, C.J., Wang, X., Keutzer, K., Darrell, T.: Region similarity
  representation learning. arXiv preprint arXiv:2103.12902  (2021)

\bibitem{xie2021detco}
Xie, E., Ding, J., Wang, W., Zhan, X., Xu, H., Li, Z., Luo, P.: Detco:
  Unsupervised contrastive learning for object detection. arXiv preprint
  arXiv:2102.04803  (2021)

\bibitem{xie2017aggregated}
Xie, S., Girshick, R., Doll{\'a}r, P., Tu, Z., He, K.: Aggregated residual
  transformations for deep neural networks. In: Proceedings of the IEEE
  conference on computer vision and pattern recognition. pp. 1492--1500 (2017)

\bibitem{xie2021propagate}
Xie, Z., Lin, Y., Zhang, Z., Cao, Y., Lin, S., Hu, H.: Propagate yourself:
  Exploring pixel-level consistency for unsupervised visual representation
  learning. In: Proceedings of the IEEE/CVF Conference on Computer Vision and
  Pattern Recognition. pp. 16684--16693 (2021)

\bibitem{zhang2016colorful}
Zhang, R., Isola, P., Efros, A.A.: Colorful image colorization. In: European
  conference on computer vision. pp. 649--666. Springer (2016)

\bibitem{zhang2017split}
Zhang, R., Isola, P., Efros, A.A.: Split-brain autoencoders: Unsupervised
  learning by cross-channel prediction. In: Proceedings of the IEEE Conference
  on Computer Vision and Pattern Recognition. pp. 1058--1067 (2017)

\end{thebibliography}
\end{document}